\documentclass{article}

     \usepackage[final,nonatbib]{neurips_2020}

\usepackage[utf8]{inputenc} 
\usepackage[T1]{fontenc}    
\usepackage{url}            
\usepackage{booktabs}       
\usepackage{amsfonts}       
\usepackage{nicefrac}       
\usepackage{microtype}      
\usepackage{epsfig}
\usepackage{graphicx}
\usepackage{color}      
\usepackage{xcolor}      
\usepackage{array}
\newcolumntype{P}[1]{>{\centering\arraybackslash}p{#1}}
\usepackage{bbm}
\usepackage{enumitem}
\usepackage{appendix}

\definecolor{citecolor}{HTML}{2980b9}
\definecolor{linkcolor}{HTML}{c0392b}
\usepackage[pagebackref=true,breaklinks=true,colorlinks,bookmarks=false,citecolor=citecolor,linkcolor=linkcolor]{hyperref}

\newcommand{\thor}{\mbox{\sc{AI2thor}}}

\title{Learning About Objects\\ by Learning to Interact with Them }

\author{Martin Lohmann\textsuperscript{1}, Jordi Salvador\textsuperscript{1}, Aniruddha Kembhavi\textsuperscript{1,2}, Roozbeh Mottaghi\textsuperscript{1,2}\\[5pt] \textsuperscript{1}PRIOR @ Allen Institute for AI \,\,\,\,\,\, \textsuperscript{2}University of Washington\\\url{https://prior.allenai.org/projects/learning_from_interaction}}

\begin{document}

\maketitle

\begin{abstract}
Much of the remarkable progress in computer vision has been focused around fully supervised learning mechanisms relying on highly curated datasets for a variety of tasks. In contrast, humans often learn about their world with little to no external supervision. Taking inspiration from infants learning from their environment through play and interaction, we present a computational framework to discover objects and learn their physical properties along this paradigm of Learning from Interaction. Our agent, when placed within the near photo-realistic and physics-enabled \thor\ environment, interacts with its world and learns about objects, their geometric extents and relative masses, without any external guidance. Our experiments reveal that this agent learns efficiently and effectively; not just for objects it has interacted with before, but also for novel instances from seen categories as well as novel object categories.
\end{abstract}

\section{Introduction}
Over the past few years, the computer vision community has witnessed remarkable breakthroughs on a variety of tasks such as image classification \cite{alexnet}, object detection \cite{fasterrcnn} and semantic segmentation \cite{deeplab}. Much of this progress can be attributed to the re-emergence of deep learning in an era of tremendous compute, and importantly, the availability of large annotated datasets that enable models to be trained in a fully supervised learning paradigm. While recent works \cite{moco,simclr} have shown an impressive ability to train visual representation stacks in a self-supervised manner, downstream applications using these representations continue to be trained with fully supervised methods \cite{goyal2019scaling}.

In stark contrast, humans often learn about their world with little or even no external supervision. For instance, infants learn about objects in their physical environment and their behaviors just by observing~\cite{Baillargeon2004InfantsPW} and interacting~\cite{Gopnik1999TheSI} with them. Inspired by these studies, we propose a computational approach to discover objects and learn their physical properties in a self-supervised setting along this paradigm of \emph{Learning from Interaction}.

Learning about objects by interacting entails the following steps: First, given an environment, the learning agent must pick a location in space, perhaps an object, to interact with. Second, the agent must determine the nature of this interaction (for instance, pushing, lifting, throwing, etc). Third, the result of this interaction must be interpreted solely via visual feedback and with no external supervision. And finally, it must iterate over these steps effectively and efficiently with the goal of learning about objects and their attributes. For instance, an agent attempting to learn to estimate the mass of objects based on their appearance may find it beneficial to interact with a diverse set of objects, pick or push them and finally acquire feedback via a combination of the force applied and the resultant displacement on the objects.

\begin{figure}[tp]
    \centering
    \includegraphics[width=32pc]{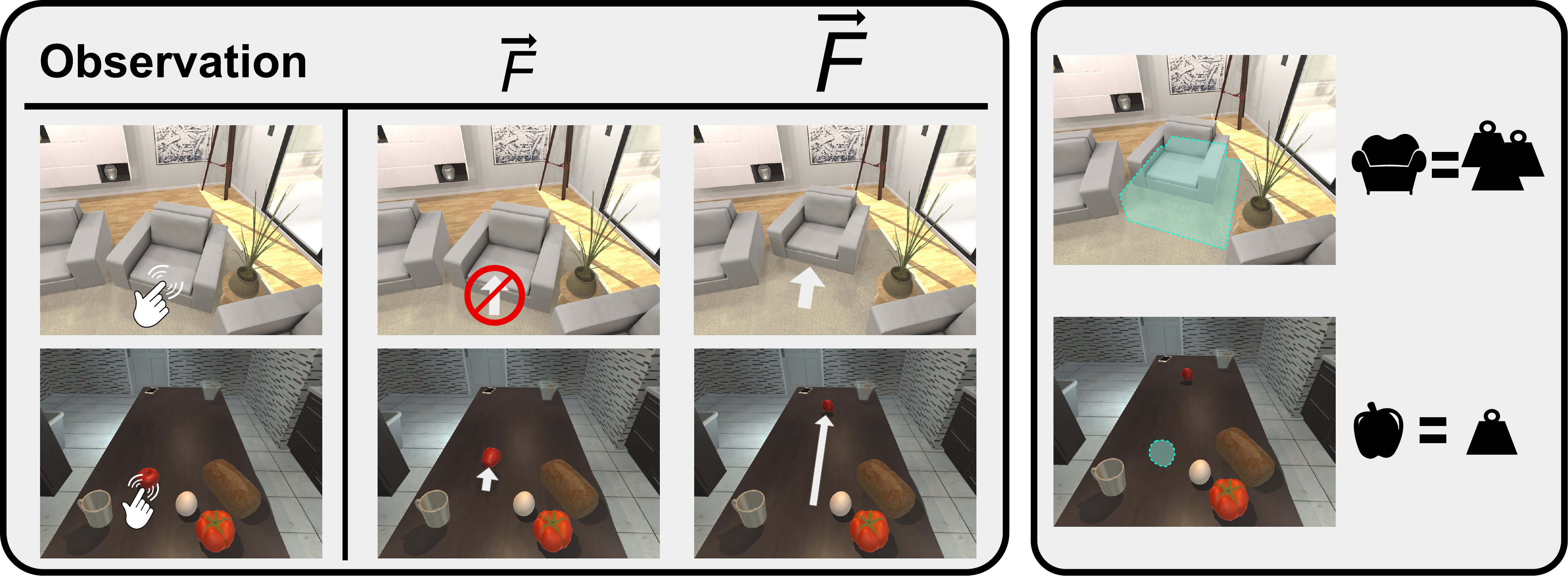}
    \caption{Our goal is to learn geometric extents (segmentation) and masses of objects in a self-supervised fashion by interacting with the surrounding world. The agent should learn not only which parts of the current observation are interactable but also how to interact with them. For example, a small force may not move a sofa but moves an apple, which enables estimating the object properties.}
    \label{fig:teaser} 
\end{figure}

In this work, we present an agent that learns to locate objects, predict their geometric extents as well as relative masses merely by interacting with its environment, with no external supervisory signal (Figure~\ref{fig:teaser}). Our proposed model learns to predict what it should interact with and what forces should be applied to those points. The interaction results in a sequence of raw observations, which are used as supervision for training a CNN model. The raw observations enable computing self-supervised losses for the interaction point, the amount of force, and for clustering points that move coherently upon interaction and so probably belong to a single object. To stabilize training and make it more efficient, we use a memory bank with prioritized sampling which is inspired by prioritized replay and self-paced learning. 

We train and evaluate our agent within the \thor~\cite{ai2thor} environment, a near photo-realistic virtual indoor environment of 120 rooms such as kitchens and living rooms with more than 2,000 object instances across 125 categories. Importantly, \thor\ is based on a physics engine which enables objects to have physical properties such as mass, friction and elasticity and for objects and scenes to interact realistically with each other. Our agent is placed in this world with no prior knowledge (apart from the inductive bias of a CNN and a self-supervision module), interacts with this world by applying forces, starts learning about the presence of objects via their displacement and eventually learns to visually estimate their attributes. Experimental evaluations show that our model obtains promising results for novel instances of seen object categories as well as unseen object categories.
\section{Related Work}
We now present a series of works that explore the problems of object discovery and mass estimation, primarily using self-supervised learning mechanisms, including interaction.

\noindent\textbf{Segmentation by interaction.} A large number of past works in the robotics community have explored the problem of segmenting objects by interacting with them~\cite{fitzpatrick03, kenney09,bjorkman10,nalpantidis12,vanHoof14,hausman15,pajarinen15,byravan17,eitel19}. These works typically use physical robots requiring them to also overcome low level manipulation challenges. Due to the size and anchored nature of these robots, experiments are typically carried out in constrained laboratory settings, which may not generalize to real world scenarios with complex backgrounds. While our experiments are carried out in simulation, \thor\ provides a large number of scenes with varied backgrounds and objects. \cite{pathak18} address instance segmentation by interaction, also dealing with physical robots and simple backgrounds. It assumes that objects go out of the field of view after the interaction and the images are captured in a fixed setup by multiple cameras, simplifying ground truth estimation. \cite{agrawal16} learn an intuitive model of physics by poking objects and \cite{pinto16} identify grasp locations using self-supervision. Our approach addresses orthogonal problems of discovering objects and learning attributes. 

\noindent\textbf{Object Proposals.} The task of identifying all objects in an image is sometimes referred to as object proposal generation, and is an essential component of many state-of-the-art object detectors \cite{fasterrcnn, maskrcnn}. Both supervised (e.g., \cite{zitnick14,kuo15}) and weakly supervised (e.g., \cite{tang18}) object proposal generation approaches have been developed. In this paper, we tackle a similar problem, but we rely on interaction and self-supervision to find objects. Past works have explored unsupervised approaches for object proposal generation from single images. \cite{deng17} propose an approach for finding 3D proposals in RGBD images using geometric means. Unlike our approach they do not use motion cues in an interactive environment. \cite{eslami16} propose a generative model to find objects in an image. However, their approach does not scale to large images or images with multiple objects. 

\noindent\textbf{Multi-frame Object Discovery.} A large body of work addresses the problem of finding objects in multiple images or videos. The supervised video object discovery approaches use different forms of supervision such as CNNs pre-trained for other tasks \cite{liang15,tsai16,tokmakov17}, supervised object proposal models \cite{fragkiadaki15,cho15,kwak15}, fixation data as annotation \cite{wang19}, or instance segmentation for single frames \cite{perazzi17}. Some of these approaches are considered unsupervised in that there is no supervision across the video frames. In contrast, our method is self-supervised and we do not rely on any annotation. 

Unsupervised and semi-supervised object discovery approaches have been explored as well \cite{rubinstein13,siva13,ochs14,misra15,stretcu15,xiao2016,croitoru17,caelles17, wang19zero,lu20}. \cite{ochs14,misra15,caelles17} require annotations for the first or few frames of the video. In contrast, our method requires no annotation. \cite{stretcu15,xiao2016,wang19zero} require a video at test time to find the moving objects. Our approach infers the objects from a single image. \cite{rubinstein13} propose an unsupervised approach to discover objects in a collection of web images. In contrast, we identify objects without any prior knowledge of objects in images. 
\cite{lu20} propose a multi-granular approach to find objects in videos. \cite{siva13} find object proposals by sampling a saliency map. \cite{croitoru17} also propose an unsupervised approach, which uses PCA reconstruction. Note that all these approaches typically assume the object is moving. This assumption is not valid in general, especially in indoor environments, since the objects are typically at rest. Hence, our agent depends on its actions to set objects in motion. 

\noindent\textbf{Mass Estimation.} Estimating mass from visual data has been explored in different contexts. \cite{wu15} infer the mass by observing objects sliding down a ramp. \cite{Standley2017image2massET} infer the mass in a supervised setting. There are also approaches for mass estimation for specific object categories (e.g., \cite{Pfitzner2015Libra3DBW,Aujeszky2019EstimatingWO}). Our work differs from these works since we infer the relative mass of objects by interacting with them.

\section{Task: Self-Supervised Object Attribute Estimation}

Humans typically learn about objects in their vicinity by observing them as well as interacting with them. There is  little to no external supervision given by other humans. We present a computational approach to discover objects and estimate their attributes in a self-supervised fashion, using no external supervision.

Our agent does not interact with the physical world, but instead, is instantiated within \thor~\cite{ai2thor}, a virtual near photo-realistic environment with a physics engine, that is often used to study embodied agents \cite{Huang2019NeuralTG, Gan2019LookLA, Jain2019TwoBP, khz2020visualreaction, Gordon2017IQAVQ}. \thor\ provides a large number of indoor scenes within which agents may navigate around, reach for objects, apply forces to them, pick and throw them. Hence, it is a suitable testbed for our interactive agent to learn about its world.
In this work, we attempt to discover objects with no supervision and estimate two attributes - geometric extent (via a 2-d segmentation of the observed pixels) and an appropriate force to move the object, which we refer to as relative mass. At each training episode, the agent is spawned at a random location within one of the scenes, and observes an ego-centric RGB+Depth view. It then actively generates raw visual feedback by picking $N$ points $(u_i,v_i)_{1:N}$ within the image to interact with, each with an interaction force magnitude $(f_i)$. These forces are applied sequentially in time, to the points $(u_i, v_i)$. The force is applied to any object or structure within the reach of the agent along the ray originating from the center of the camera and passing through that point. If the point $(u_i,v_i)$ does not correspond to any movable object or the force pushes an object against static obstacles, there is typically little to no perceived change in the observation\footnote{Small changes may be observed due to fluctuations in lighting.}, but if it corresponds to an object and the chosen force is sufficiently strong, the object moves. The agent then receives raw visual feedback consisting of the RGB view after the scene has come to rest. Through a sequence of interactions within the training scenes, the agent must learn to (1) identify points $P$ in an observation that are likely to cause motion, corresponding to interactable objects and (2) estimate their attributes. During evaluation, the agent must predict these without interacting with the scene, instead using only a single visual observation as input.

Self-supervised object discovery with attribute estimation poses several challenges: 
(1) \emph{Sparse supervision:} Supervised learning frameworks typically provide dense annotations. For example, instance segmentation datasets provide mask annotations for all object instances within each target category. In contrast, our agent obtains supervision only for points that it chooses to interact with. The remaining points cannot be safely regarded as background since a scene may have unexplored objects within it. The small number of interactions per scene (owing to a fixed budget) and their sequential nature (so that scene changes accumulate over time) prevent the agent from obtaining dense supervision. 
(2) \emph{Noisy supervision:} In our setting, supervision must be computed from the observations before and after interaction and cannot utilize a trained surrogate model and thus tends to be very noisy. In addition, some of the supervision is inherently ambiguous. For instance, small movements can only help segment an object partially. 
(3) \emph{Class imbalance:} In a typical household scene, objects that can be moved by a force occupy a tiny fraction of the total room volume. This causes a high imbalance between object and non-object pixels, which complicates learning. 
(4) \emph{Efficiency:} While significantly cheaper than using a physical robot, interacting with a virtual environment is time consuming, compared to methods dealing with static images. Our agent is assigned a fixed budget of interactions and must learn to use this wisely.

\section{Model}
\label{sec:model}

\begin{figure}[tp]
    \centering
    \includegraphics[width=33pc]{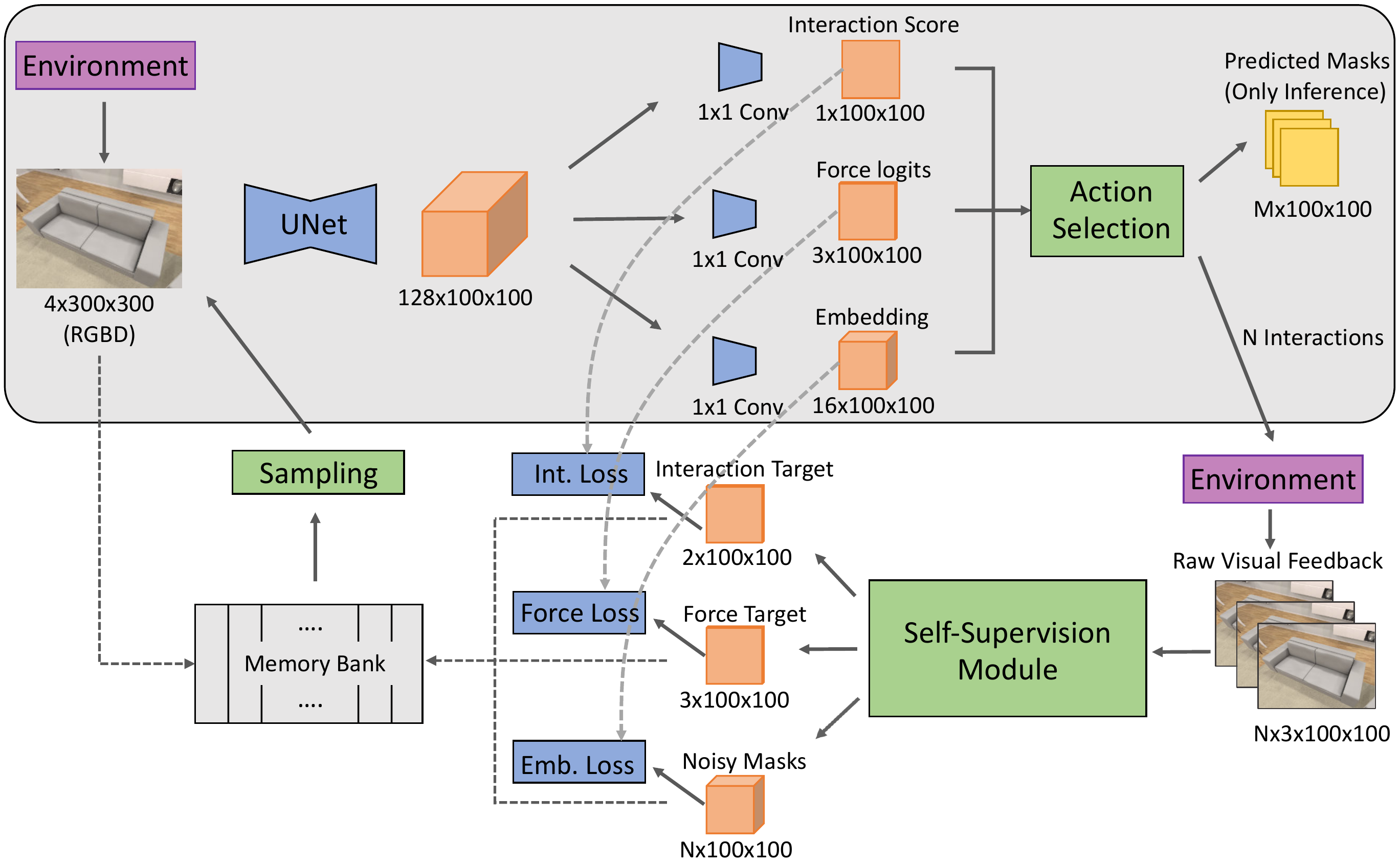}
    \caption{\textbf{Model overview}. Our model is a convolutional neural network that receives an RGB+D input and outputs instance masks and relative mass estimates. The grey box shows the portion of the network used during inference. The outputs of each network component are shown in orange. We compute three types of losses: Interaction loss, Force loss and Embedding loss. Note that both the predictions and, via interaction, the ground truth are generated by the network.}
    \label{fig:model} 
\end{figure}

Our model design is a convolutional neural network inspired by past works in clustering based instance segmentation \cite{neven19,fathi17}. As shown in Figure~\ref{fig:model}, it inputs a single $300\times300$ RGB+D image and passes it through a UNet style backbone \cite{unet}, which also consumes absolute pixel coordinates similar to \cite{liu18}. The network produces three output tensors (shown in orange in the top portion of Figure~\ref{fig:model}), each with a $100\times100$ spatial extent: (1) An interaction score per location - This signifies the confidence of the model that an interaction with this pixel will result in a change in its observation. (2) Force logits per pixel - These indicate the minimum force magnitude the model predicts will be necessary to achieve such a change. In practice, we quantize force magnitudes into 3 bins, resulting in 3 logits per location. (3) Spatial embeddings - These are computed for each spatial location and capture the appearance of that location. The embeddings are used in a clustering algorithm to compute object instance segmentation masks, which encourages locations within a single object to have similar embeddings, and locations across objects to have different embeddings. Each output is trained with its own loss function. These tensors are used to select actions during training and estimate object attributes at inference.

The network has a field of view large enough to include several objects but small enough to ensure sufficient resolution for small objects within a $300 \times 300$ image. Our network has relatively small number of parameters (1.4M), which helps stabilize training. See Appendix~\ref{app:model} for more details.

\subsection{Inference}
\label{sec:inference}
At inference, the model predicts points and force magnitudes of interaction, and a binary instance proposal segmentation mask corresponding to each point, via the following algorithm. During training, the agent uses the same inference procedure to select a set of actions (locations $(u_i,v_i)_{1:N}$ and forces $(f_i)_{1:N}$). \\ 
\textbf{Given}: A network forward pass, desired number of actions ($N$) and interaction score threshold ($\theta$).\\
\textbf{Iterate} until $N$ actions selected or all interaction scores $< \theta $:
\begin{enumerate}[noitemsep,topsep=0pt,parsep=0pt,partopsep=0pt]
    \item Select the pixel $\mathbf p = (u,v) $ with the highest interaction score. Select the force magnitude $f^r$ (details in Sec~\ref{sec:training}-II) with $r = \mathrm{argmax}_r \, m^r_{\mathbf p} $, where $(m^r_{\mathbf p})_{r=1:3} $ denotes the vector of force logits at $\mathbf p$. Further, denote by $e_{\mathbf p}$ the $16D$ embedding vector of this pixel.
    \item Add the pixel $\mathbf p $ and $f^r$ to the list $(u_i,v_i)_{1:N}$ of action points and $(f_i)_{1:N}$ of forces.
    \item Consider the set $\tilde{P}$ of pixels $\mathbf p' $ such that\footnote{Threshold $1$ is determined by the scale of the Embedding Loss. See Appendix~\ref{sec:losses}.} $\Vert e_{\mathbf p'} - e_{\mathbf p}\Vert_2 < 1 $. Define $e$ to be the $16D$ mean vector of all $e_{\mathbf p'},\, \mathbf p'\in \tilde{P} $.
    \item Consider the set $P_M $ of pixels $\mathbf p'$ such that $\Vert e_{\mathbf p'} - e\Vert_2 < 1 $. Add this set to the list of proposed object segmentation masks.
    \item Set interaction scores and embedding vectors to $-\infty $ for all $\mathbf p' \in P_M $.
\end{enumerate}

Note that the relative mass (predicted force magnitude) of an object is the argmax of the force logits at the pixel that the model chose for interaction with the object. While we predict such force logits for every pixel in the image, they are used only on points selected for interaction.

Intuitively, each iteration in the algorithm above runs one step of an EM clustering algorithm with seed selected by the interaction score, and an $L^2$ based similarity metric for the embedding vectors. Step 1 chooses the maximum interaction score among all pixels that have not yet been clustered as part of an object, and so the algorithm is greedy. During self-supervised training, we further add random actions to the actions selected by the algorithm, to aid exploration.

\subsection{Training}
\label{sec:training}
Training the network using only self supervision proceeds iteratively as follows:

\textbf{I: Action selection.} The agent is spawned at a random location within the training scenes. The model receives an RGB+D image $I_1$ as its observation, passes it through the network and selects interaction locations and corresponding forces. This is identical to the inference procedure outlined above.

\textbf{II: Interaction.} The agent obtains a sequence of raw visual feedback by applying forces to the chosen interaction points in the order determined by the model. Let $F = \{f^0,f^1,f^2\}$ be a set of force magnitudes that are chosen \emph{a priori} and correspond to magnitudes typically required to move light, medium and heavy objects\footnote{We have selected typical minimal forces that move objects which correlate strongly, but not perfectly, with true physical masses. Our model learns such forces, and we evaluate against (quantized) ground truth masses.}. Let $D$ be a set of 8 force directions corresponding to a quantized set of directions in space, but avoiding those that point into the ground. Finally, let $r$ be the predicted relative mass of the object (if any) at the interaction point ($r=0,1,2$ $\rightarrow$ light, medium, heavy).

To test whether the chosen force magnitude $f^r$ is too large for the object proposal, the agent first applies a force with magnitude $f^{r-1}$ to the chosen point, with force direction randomly drawn from $D$. If no change is detected in the visual feedback, it increases the force to the predicted magnitude $f^{r} $. If again no changes occur, it applies the maximum force magnitude ($f^2 $ in our case), which by definition will move any un-obstructed, interactable object if one were present at the chosen point. This provides noisy supervision that helps the agent learn whether the predicted force was too large, just right, or too low. Note that the agent might apply a force in an undesirable direction resulting in no change (e.g., pushing an object against a wall or other obstacles). Such issues introduce further noise into the supervision signal, but can't be avoided in realistic scenarios.

\textbf{III: Self supervision.} Our goal is to only use self-supervisory signals to train the network. Hence, we rely on simple visual changes between the observations, prior to and after the application of the force. We detect changes in the scene using image difference. Consider a pair $I_k,\, I_{k+1} $ of consecutive views of the scene, before and after interaction, transformed to HSV space. We downsample the image difference $J=I_{k}-I_{k+1}$ using mean pooling to have the same size as the output of the forward pass. We compute a binary mask $B=\mathbbm{1}(J^2>0.01)$, where $J^2$ is the pixel-wise $L^2$ norm. 

To alleviate some of the noise introduced into the supervision and to align the self-supervised mask along the edges of objects, we employ an unsupervised low-level grouping of pixels to form superpixels \cite{felzenszwalb2004efficient} and use them to post-process the mask $B$ into a more robust mask $B^+$, which is a union of all superpixels that overlap sufficiently with the noisy mask $B$. More precisely, we compute superpixels for image $I_1$ (the original image received by the model). For each superpixel, if at least 25\% of its pixels belong to $B$, we add the superpixel to the new mask $B^+$. If the number of pixels on $B^+$ in the immediate vicinity of the interaction location, weighted inversely by distance, is less than a threshold, we declare that the scene has not changed and consider the interaction ``unsuccessful''. Otherwise, we consider it ``successful'', and consider $B^+$ as the (noisy) segmentation mask of the object that moved during the interaction. While this method is simple and completely unsupervised, it is noisy due to imperfect thresholding, appearance challenges such as change in shadows, movements of multiple objects via a single interaction, object state changes (e.g., a high force might break a glass), partial overlap between object masks in the two frames, etc.

Finally, the initial frame $I_1$, successful interaction points, and associated supervision masks $B^+$, predicted force magnitudes, and the true force magnitudes that produced the change for these interactions, as well as the points leading to unsuccessful interactions are added to a memory bank.

\textbf{IV: Learning from the memory bank.} After a number of interactions, gradient updates are performed on batches sampled from the memory bank based on an importance score. We have noticed in experiments that under-sampling medium-loss training data yields substantial performance gains, possibly because fitting the noise patterns in such data would be detrimental to the learning process. This should be compared to prioritized replay methods \cite{schaul16} (where high-loss, difficult, images are over-sampled) and self-paced learning \cite{kumar10,bengio09} (where low-loss, easy images are over-sampled). More precisely, data points that are yet to take part in an update are assigned a high importance score and data points with no detected objects (in step III) are assigned a low score. When a data point takes part in an update, its score is reassigned based on the IoU of its predictions with the noisy segmentation targets of the memory bank. Details are in Appendix~\ref{app:memory}.

For each batch of image-target pairs added to the memory bank, we sample $K$ batches from the bank and backpropagate the losses on these batches. $K$ is annealed during training. Interacting with a simulated world, while not as slow as a real robot, still consumes a fair bit of time due to invoking the physics engine. A memory bank allows every interaction to be sampled multiple times over training, making our model more robust and also more efficient than a pure online approach. We use a fixed size memory bank (20,000 points). Once filled, new points replace the oldest ones. After gradient updates for each batch, the agent is spawned at new locations and repeats this process. 

\subsection{Loss Functions}

We now describe the loss functions used to learn the interaction score, force logits, and the pixel-wise spatial embedding vectors. Recall that for each scene the agent has previously interacted with, the memory bank contains successful and unsuccessful locations of interaction, and also force magnitudes and noisy masks for each successful interaction.

\textbf{Interaction score loss.} We want to yield high interaction scores at the points of successful interactions, low scores at points of unsuccessful interactions, but do not supply the model with gradients from unexplored regions of the image (where no interactions were performed). We first form two targets: a binary map for foreground locations and one for background locations. These are smoothed via a Gaussian kernel and then used in a KL divergence-based loss, but with gradients that decay faster with model confidence, similar to a focal loss \cite{focalloss}.

\textbf{Force loss.} For each successful interaction, the memory bank stores if the predicted force magnitude $f^r $ was (1) correct, (2) too small to move the object, or (3) too large. The force loss penalizes cases 2 and 3. In case 1, we want to increase the $r$th force logit at this position. In case 2, we provide gradients that increase all force logits $s $ with $ s>r$. In case 3, gradients increase all force logits $s $ with $s<r$. Pixel-wise force logits only receive gradients at successful interaction locations. 

\textbf{Embedding loss.} The inference method outlined in Sec~\ref{sec:inference} employs a clustering method over embedding vectors to determine instance segments. Our goal is to decrease the variance among embedding vectors that belong to the same object and simultaneously increase the dissimilarity of embedding vectors belonging to different objects. We use the noisy masks produced by the self-supervision module to supervise the embedding vectors. For each mask, we compute the mean embedding vector $e$ of all pixels on the mask, and compute $\Vert e_{\mathbf p} - e\Vert_2$ for all pixels. We provide gradients that decrease this distance for all $e_{\mathbf p} $ with $\mathbf p$ belonging to the mask, and increase the distance for all $\mathbf p $ not on the mask. To stabilize training, inspired by the focal loss, our loss has gradients that decrease rapidly for confident predictions (very large or very small $\Vert e_{\mathbf p} - e\Vert_2 $). 

The primary advantage of these losses, and a clustering approach to segmentation, is that no gradients are associated to unexplored regions of the image, addressing one of our key challenges above. The losses are robust to noise, and sparse enough to be partially shielded from class imbalance between object and non-object pixels, addressing the other important challenges.

\section{Experiments}
We now describe experiments and ablations on our self-supervised method for instance segmentation and mass estimation. As a point of reference, we also include results for different levels of supervision.

\begin{table*}[t!]\footnotesize
  \centering
\begin{small}
\centerline{
    \begin{tabular}{l cc cc}
    \toprule
    &\multicolumn{2}{c}{\centering{Bounding Box}} & \multicolumn{2}{c}{\centering{Mask}} \\
        & AP$^{\scriptsize{\mbox{IOU=0.5}}}$ & AP & AP$^{\scriptsize{\mbox{IOU=0.5}}}$ & AP \\
    \midrule
    \multicolumn{5}{l}{\bf Joint segmentation \& mass:} \\
    (a) Ours (self supervision) & 24.19 / 27.59 & 11.65 / 13.44  & 22.00 / 25.01 & 10.24 / 11.00 \\
    (b) Mask-RCNN \cite{maskrcnn} (full supervision) & 33.94 / 50.72 & 22.08 / 34.44 & 28.21 / 45.57 & 16.75 / 30.12 \\
    \midrule
    \multicolumn{5}{l}{\bf Segmentation only:} \\
    (c) Ours (self supervision) & 27.06 / 26.10 & 12.71 / 12.05 & 24.37 / 23.70 & 11.09 / 10.30 \\
    (d) Ours w/ supervised masks & 30.77 / 38.44 & 16.83 / 21.25 & 27.46 / 36.49 & 13.89 / 18.67 \\
    (e) Ours w/ supervised interaction & 26.11 / 28.06 & 12.26 / 13.23 & 24.87 / 26.14 & 11.53 / 11.71 \\
    (f) Ours (full supervision) & 30.23 / 25.68 & 14.86 / 12.03 & 27.40 / 24.54 & 13.23 / 11.00 \\
    \midrule
    (g) VideoPCA \cite{croitoru17} (supervised interaction) & 8.10 / 7.48 & 3.99 / 3.45 & 7.33 / 6.97 & 3.27 / 2.82\\
    (h) Mask-RCNN \cite{maskrcnn}  (full supervision) & 36.49 / 49.86 & 22.92 / 31.63 & 30.66 / 44.87 & 18.35 / 28.85 \\
    (i) Mask-RCNN \cite{maskrcnn} (self-supervised masks)  & 15.93 / 21.26 & 8.16 / 11.39 & 10.88 / 15.84 & 6.08 / 8.73 \\
    (j) Robust Set Loss \cite{pathak18} (self-supervised masks) & 17.51 / 22.29 & 9.11 / 12.08 & 10.71 / 16.23 & 4.02 / 8.81 \\
    \midrule
    \multicolumn{5}{l}{\emph{\textbf{Ablations (segmentation only)}}} \\
    (k) Ours w/o superpixels & 12.74 / 13.24 & 4.77 / 4.93 & 10.21 / 10.72 & 3.81 / 3.75 \\
    (l) Ours w/o prioritized sampling & 21.04 / 21.32 & 8.70 / 8.78 & 19.02 / 19.68 & 7.75 / 7.76 \\
    (m) Ours $\infty$ arm length & 21.06 / 20.10 & 9.38 / 9.36 & 20.64 / 17.26 & 9.15 / 7.56 \\
    \bottomrule
  \end{tabular}
  }
  \end{small}
  \caption{\textbf{Object segmentation.} Results are for \emph{NovelObjects} / \emph{NovelSpaces} scenarios.}
  \label{tab:results}
\end{table*}

\noindent \textbf{Framework.} Objects and interactions in \thor\ \cite{ai2thor} are governed by an underlying Unity physics engine. This enables objects to have fairly realistic physical attributes like mass, material and friction. \thor\ contains four types of scenes: \emph{kitchens}, \emph{living rooms}, \emph{bedrooms}, and \emph{bathrooms}, with 30 rooms of each type. We conduct experiments with two data splits: (1) \emph{NovelSpaces} - We use 80 scenes for training (20 in each type) and report results on 20 different rooms (5 in each type). (2) \emph{NovelObjects} - We use kitchens and living rooms in train (60 scenes) and bedrooms (30 scenes) in test. In this scenario, the majority of object categories encountered in test are novel (e.g. pillows are only seen in bedrooms). In addition, object locations within each scene in \thor\ are also randomized. This provides ample variability during training.

In total, there are 17,211 locations with 2,765 interactable object instances within the reach of the agent split across scenes in both data splits. \thor\ includes 125 object categories in total, where 86 of them are interactable. Note that, in many locations, there are no objects in reach of the agent to interact with, and it should learn no interactable objects exist in those locations. \\
We train all networks from scratch. Implementation details are provided in Appendix~\ref{sec:implementation}.

\noindent \textbf{Object segmentation.}  Table~\ref{tab:results} shows results for segmenting objects. We report standard COCO metrics \cite{coco} for detection and instance segmentation. Row (a) shows results for a model that predicts segments and mass jointly. This shows fairly strong results, given no annotations during training, even in the harder NovelObjects scenario; illustrating our approach generalizes well to object categories that have never been seen during training. (c) shows results for a model trained only for segmentation. It is interesting that in spite of solving a harder task, (a)'s metrics do not reduce a lot compared to (c). 

\noindent \textbf{Mass estimation.} We evaluate the model for mass estimation using three buckets: mass $m<0.5$ kg; $0.5\leq m<2$ kg; and $m\geq 2$ kg with results in Table~\ref{tab:massres}. The reported metric is mean per class accuracy, for which chance is at 33\%. We also report AP in which case a bounding box is considered correct if both the extent and the mass is predicted correctly. This is a strict metric and, as can be expected, the model shows low scores.  For self-supervision, we fix the set of forces to $f^0=5$N, $f^1=30$N and $f^2=200$N in order to provide reasonable interactions with objects in each bucket. 

\textbf{Providing more supervision.} While our method has been developed for self-supervised learning from noisy feedback, we also explore how supplying ground truth information aides the learning process. For these, we consider only the pure instance segmentation model within Table~\ref{tab:results}. (e)-\textit{Supervised interactions} shows improved results for NovelSpaces, when an oracle supplements interaction locations for all objects in the scene not selected by the model. (d)-\textit{Supervised masks} shows results when an oracle provides the true segmentation masks of all objects selected for interaction by the model. This shows a huge improvement over (c), reflecting the difficulty of learning from noisy masks in a self-supervised setting. Finally, (f)-\textit{Full supervision} combines both levels of supervision. Surprisingly, this does not perform much better than (c). Qualitatively, we notice that (f) over-segments objects and is usually over-confident about them. While this behavior may be due to the fact that our model and training design have been developed for the noisy self-supervised case, we also conjecture that self-supervision and noise tend to focus the attention of learning on ``easier'' objects (such as ones that move more consistently and noise-free when interacted with), which provides a natural pace for the learning progress of the model.

As a reference, (h) reports the performance of a ResNet-18 based Mask-RCNN \cite{maskrcnn} with RGB+D input trained with standard, full supervision. We use all standard settings such as optimizers, learning rate schedules, etc. Since the ResNet-18 has about 7 times more parameters than our backbone, this precludes a direct comparison to (f), but provides a useful point of reference. In (i) and (j), we report the performance of this model trained on self-supervised masks obtained from ``oracle'' interaction as described above, using pixel-wise cross-entropy loss in (i) and the robust set loss from \cite{pathak18} in (j). In \cite{pathak18}, this loss was proposed to deal with noisy segmentation masks in their self-supervised learning scenario. The performance of both losses in our case is similar. The massive drop from (h) once again indicates the difficulty of learning from noisy masks.

\begin{figure}[tp]
    \centering
    \includegraphics[width=33pc]{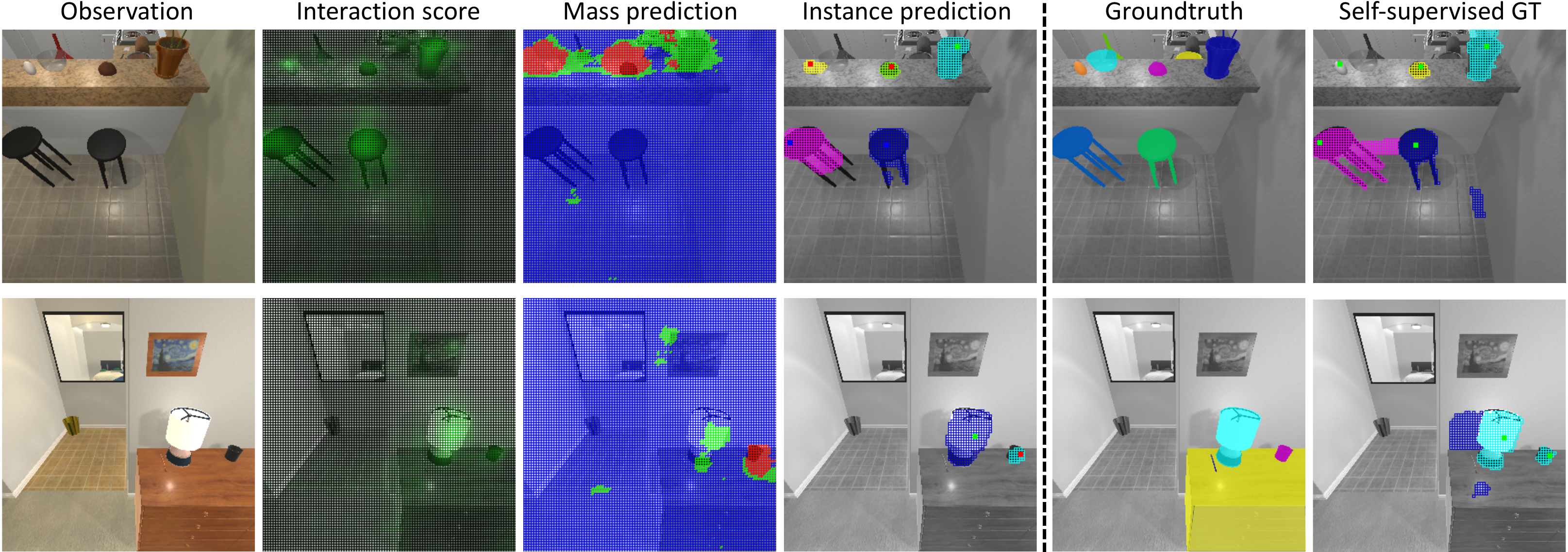}
    \caption{\textbf{Qualitative results.} Instance segmentation and mass prediction results are illustrated. The brighter green corresponds to higher interaction score. The masses are shown in red (light), green (medium) and blue (heavy). The selected interaction points are shown with colored dots.}
    \label{fig:qual} 
\end{figure}

\begin{table*}[tp]\footnotesize
  \centering
\begin{small}
    \begin{tabular}{l cc}
    \toprule
        & Mean per Class Acc. & AP$^{\scriptsize{\mbox{IOU=0.5}}}$ (Mass \& BBox)  \\
    \midrule
    Ours (self-supervised) & 50.79 / 55.86 & 11.85 / 11.01 \\
    Mask-RCNN (full supervision) & 78.28 / 86.11 & 26.90 / 46.24 \\
    \bottomrule
  \end{tabular}
  \end{small}
  \caption{\textbf{Mass estimation.} Results are for \emph{NovelObjects} / \emph{NovelSpaces} scenarios.}
  \label{tab:massres}
\end{table*}

\textbf{Other types of self supervision.} We also provide the result of \cite{croitoru17} which extracts segmentation masks using a principal component analysis of the observation sequence as a result of the agent's interaction (row (g)). This method was not effective in discovering objects. We also used an unsupervised optical flow method \cite{meister18} to compute supervision masks, which did not work well either. These results highlight the difficulty of signal extraction from raw visual feedback.

\noindent \textbf{Ablations.} We first ablate the effect of the visual prior built into our self-supervision module - using super-pixels as a post-processing step to obtain supervision masks (row (k)), resulting in a large drop. Next, we ablate the effect of our memory bank sampling procedure (row (l)), which also shows a large drop, validating our design choices. Row (m) shows the effect of increasing the agent's interaction radius; increasing the number of objects used for evaluation, and rendering the task harder.

\noindent \textbf{Qualitative results.} Figure~\ref{fig:qual} shows some qualitative results of our self-supervised approach used to estimate object extents (instance segmentation) and masses. Our model can recognize multiple objects even in cluttered scenes, and for many object types produces relatively accurate masks. The method has difficulty detecting tiny objects such as pencils or forks (which need high precision for successful interaction) and large objects such as sofas or drawers (which move only slightly during interaction, and whose movement is usually detected only at the boundary of the object). Predicting masses from visual cues is expectedly a hard problem as seen by our quantitative and qualitative results, but our model does reasonably well to distinguish `light' and `medium' from `heavy' objects.

\section{Conclusion}

An important component in visual learning and reasoning is the ability to learn from interaction with the world. This is in contrast to the most popular approaches to the computational models of vision that rely on highly curated datasets with extensive annotations. In this paper, we present an agent that learns to locate objects and predict their geometric extents and relative masses merely by interacting with its environment. Our experiments show that, in fact, our model obtains promising results in estimating these object attributes without any external annotation even for object categories that are novel and not observed before. Our future work involves inferring more complex object attributes such as different states of objects, friction forces and material properties. 


{\small
\bibliographystyle{ieee}
\bibliography{egbib}

\begin{thebibliography}{10}\itemsep=-1pt

\bibitem{coco}
Coco dataset.
\newblock \url{http://cocodataset.org/}.

\bibitem{agrawal16}
Pulkit Agrawal, Ashvin~V Nair, Pieter Abbeel, Jitendra Malik, and Sergey
  Levine.
\newblock Learning to poke by poking: Experiential learning of intuitive
  physics.
\newblock In {\em NeurIPS}, 2016.

\bibitem{Aujeszky2019EstimatingWO}
Tam{\'a}s Aujeszky, Georgios Korres, Mohamad Eid, and Farshad Khorrami.
\newblock Estimating weight of unknown objects using active thermography.
\newblock {\em Robotics}, 2019.

\bibitem{Baillargeon2004InfantsPW}
Renée Baillargeon.
\newblock Infants' physical world.
\newblock {\em Current Directions in Psychological Science}, 2004.

\bibitem{bengio09}
Yoshua Bengio, J\'{e}r\^{o}me Louradour, Ronan Collobert, and Jason Weston.
\newblock Curriculum learning.
\newblock In {\em ICML}, 2009.

\bibitem{bjorkman10}
M{\aa}rten Bj{\"o}rkman and Danica Kragic.
\newblock Active 3d scene segmentation and detection of unknown objects.
\newblock In {\em ICRA}, 2010.

\bibitem{byravan17}
Arunkumar Byravan and Dieter Fox.
\newblock Se3-nets: Learning rigid body motion using deep neural networks.
\newblock In {\em ICRA}, 2017.

\bibitem{caelles17}
Sergi Caelles, Kevis-Kokitsi Maninis, Jordi Pont-Tuset, Laura Leal-Taix\'e,
  Daniel Cremers, and Luc {Van Gool}.
\newblock One-shot video object segmentation.
\newblock In {\em CVPR}, 2017.

\bibitem{deeplab}
Liang-Chieh Chen, George Papandreou, Iasonas Kokkinos, Kevin Murphy, and
  Alan~L. Yuille.
\newblock Deeplab: Semantic image segmentation with deep convolutional nets,
  atrous convolution, and fully connected crfs.
\newblock {\em TPAMI}, 2018.

\bibitem{simclr}
Ting Chen, Simon Kornblith, Mohammad Norouzi, and Geoffrey Hinton.
\newblock A simple framework for contrastive learning of visual
  representations.
\newblock {\em arXiv}, 2020.

\bibitem{cho15}
Minsu Cho, Suha Kwak, Cordelia Schmid, and Jean Ponce.
\newblock Unsupervised object discovery and localization in the wild:
  Part-based matching with bottom-up region proposals.
\newblock In {\em CVPR}, 2015.

\bibitem{croitoru17}
Ioana Croitoru, Simion-Vlad Bogolin, and Marius Leordeanu.
\newblock Unsupervised learning from video to detect foreground objects in
  single images.
\newblock In {\em ICCV}, 2017.

\bibitem{deng17}
Zhuo Deng, Sinisa Todorovic, and Longin~Jan Latecki.
\newblock Unsupervised object region proposals for rgb-d indoor scenes.
\newblock {\em CVIU}, 2017.

\bibitem{eitel19}
Andreas Eitel, Nico Hauff, and Wolfram Burgard.
\newblock Self-supervised transfer learning for instance segmentation through
  physical interaction.
\newblock In {\em IROS}, 2019.

\bibitem{eslami16}
S.~M.~Ali Eslami, Nicolas Heess, Theophane Weber, Yuval Tassa, David
  Szepesvari, koray kavukcuoglu, and Geoffrey~E Hinton.
\newblock Attend, infer, repeat: Fast scene understanding with generative
  models.
\newblock In {\em NeurIPS}, 2016.

\bibitem{fathi17}
Alireza Fathi, Zbigniew Wojna, Vivek Rathod, Peng Wang, Hyun~Oh Song, Sergio
  Guadarrama, and Kevin~P. Murphy.
\newblock Semantic instance segmentation via deep metric learning.
\newblock {\em arXiv}, 2017.

\bibitem{felzenszwalb2004efficient}
Pedro~F. Felzenszwalb and Daniel~P. Huttenlocher.
\newblock Efficient graph-based image segmentation.
\newblock {\em IJCV}, 2004.

\bibitem{fitzpatrick03}
Paul~M. Fitzpatrick.
\newblock First contact: an active vision approach to segmentation.
\newblock In {\em IROS}, 2003.

\bibitem{fragkiadaki15}
Katerina Fragkiadaki, Pablo Arbelaez, Panna Felsen, and Jitendra Malik.
\newblock Learning to segment moving objects in videos.
\newblock In {\em CVPR}, 2015.

\bibitem{Gan2019LookLA}
Chuang Gan, Yiwei Zhang, Jiajun Wu, Boqing Gong, and Joshua~B. Tenenbaum.
\newblock Look, listen, and act: Towards audio-visual embodied navigation.
\newblock In {\em ICRA}, 2020.

\bibitem{Gopnik1999TheSI}
Alison Gopnik, Andrew~N. Meltzoff, and Patricia~K. Kuhl.
\newblock {\em The scientist in the crib: Minds, brains, and how children
  learn}.
\newblock William Morrow \& Co., 1999.

\bibitem{Gordon2017IQAVQ}
Daniel Gordon, Aniruddha Kembhavi, Mohammad Rastegari, Joseph Redmon, Dieter
  Fox, and Ali Farhadi.
\newblock Iqa: Visual question answering in interactive environments.
\newblock In {\em CVPR}, 2017.

\bibitem{goyal2019scaling}
Priya Goyal, Dhruv Mahajan, Abhinav Gupta, and Ishan Misra.
\newblock Scaling and benchmarking self-supervised visual representation
  learning.
\newblock In {\em ICCV}, 2019.

\bibitem{hausman15}
Karol Hausman, Dejan Pangercic, Zolt{\'a}n-Csaba M{\'a}rton, Ferenc
  B{\'a}lint-Bencz{\'e}di, Christian Bersch, Megha Gupta, Gaurav Sukhatme, and
  Michael Beetz.
\newblock Interactive segmentation of textured and textureless objects.
\newblock {\em Handling Uncertainty and Networked Structure in Robot Control},
  2015.

\bibitem{moco}
Kaiming He, Haoqi Fan, Yuxin Wu, Saining Xie, and Ross Girshick.
\newblock Momentum contrast for unsupervised visual representation learning.
\newblock In {\em CVPR}, 2020.

\bibitem{maskrcnn}
Kaiming He, Georgia Gkioxari, Piotr Doll\'{a}r, and Ross Girshick.
\newblock {Mask R-CNN}.
\newblock In {\em ICCV}, 2017.

\bibitem{Huang2019NeuralTG}
De-An Huang, Suraj Nair, Danfei Xu, Yuke Zhu, Animesh Garg, Li Fei-Fei, Silvio
  Savarese, and Juan~Carlos Niebles.
\newblock Neural task graphs: Generalizing to unseen tasks from a single video
  demonstration.
\newblock In {\em CVPR}, 2019.

\bibitem{Jain2019TwoBP}
Unnat Jain, Luca Weihs, Eric Kolve, Mohammad Rastegari, Svetlana Lazebnik, Ali
  Farhadi, Alexander~G. Schwing, and Aniruddha Kembhavi.
\newblock Two body problem: Collaborative visual task completion.
\newblock In {\em CVPR}, 2019.

\bibitem{kenney09}
Jacqueline Kenney, Thomas Buckley, and Oliver Brock.
\newblock Interactive segmentation for manipulation in unstructured
  environments.
\newblock In {\em ICRA}, 2009.

\bibitem{ai2thor}
Eric Kolve, Roozbeh Mottaghi, Winson Han, Eli VanderBilt, Luca Weihs, Alvaro
  Herrasti, Daniel Gordon, Yuke Zhu, Abhinav Gupta, and Ali Farhadi.
\newblock {AI2-THOR: An Interactive 3D Environment for Visual AI}.
\newblock {\em arXiv}, 2017.

\bibitem{alexnet}
Alex Krizhevsky, Ilya Sutskever, and Geoffrey~E Hinton.
\newblock Imagenet classification with deep convolutional neural networks.
\newblock In {\em NeurIPS}, 2012.

\bibitem{kumar10}
M.~Pawan Kumar, Benjamin Packer, and Daphne Koller.
\newblock Self-paced learning for latent variable models.
\newblock In {\em NeurIPS}, 2010.

\bibitem{kuo15}
Weicheng Kuo, Bharath Hariharan, and Jitendra Malik.
\newblock Deepbox: Learning objectness with convolutional networks.
\newblock In {\em ICCV}, 2015.

\bibitem{kwak15}
Suha Kwak, Minsu Cho, Ivan Laptev, Jean Ponce, and Cordelia Schmid.
\newblock Unsupervised object discovery and tracking in video collections.
\newblock In {\em ICCV}, 2015.

\bibitem{liang15}
Xiaodan Liang, Si Liu, Yunchao Wei, Luoqi Liu, Liang Lin, and Shuicheng Yan.
\newblock Towards computational baby learning: A weakly-supervised approach for
  object detection.
\newblock In {\em ICCV}, 2015.

\bibitem{focalloss}
Tsung-Yi Lin, Priya Goyal, Ross~B. Girshick, Kaiming He, and Piotr Doll{\'a}r.
\newblock Focal loss for dense object detection.
\newblock In {\em ICCV}, 2017.

\bibitem{liu18}
Rosanne Liu, Joel Lehman, Piero Molino, Felipe Petroski~Such, Eric Frank, Alex
  Sergeev, and Jason Yosinski.
\newblock An intriguing failing of convolutional neural networks and the
  coordconv solution.
\newblock In {\em NeurIPS}, 2018.

\bibitem{lu20}
Xiankai Lu, Wenguan Wang, Jianbing Shen, Yu-Wing Tai, David~J. Crandall, and
  Steven C.~H. Hoi.
\newblock Learning video object segmentation from unlabeled videos.
\newblock In {\em CVPR}, 2020.

\bibitem{meister18}
Simon Meister, Junhwa Hur, and Stefan Roth.
\newblock Unflow: Unsupervised learning of optical flow with a bidirectional
  census loss.
\newblock In {\em AAAI}, 2018.

\bibitem{misra15}
Ishan Misra, Abhinav Shrivastava, and Martial Hebert.
\newblock Watch and learn: Semi-supervised learning of object detectors from
  videos.
\newblock In {\em CVPR}, 2015.

\bibitem{nalpantidis12}
Lazaros Nalpantidis, M{\aa}rten Bj{\"o}rkman, and Danica Kragic.
\newblock Yes - yet another object segmentation: Exploiting camera movement.
\newblock In {\em IROS}, 2012.

\bibitem{neven19}
Davy Neven, Bert~De Brabandere, Marc Proesmans, and Luc~Van Gool.
\newblock Instance segmentation by jointly optimizing spatial embeddings and
  clustering bandwidth.
\newblock In {\em CVPR}, 2019.

\bibitem{ochs14}
Peter Ochs, Jitendra Malik, and Thomas Brox.
\newblock Segmentation of moving objects by long term video analysis.
\newblock {\em TPAMI}, 2014.

\bibitem{pajarinen15}
Joni Pajarinen and Ville Kyrki.
\newblock Decision making under uncertain segmentations.
\newblock In {\em ICRA}, 2015.

\bibitem{pathak18}
Deepak Pathak, Yide Shentu, Dian Chen, Pulkit Agrawal, Trevor Darrell, Sergey
  Levine, and Jitendra Malik.
\newblock Learning instance segmentation by interaction.
\newblock In {\em CVPR Workshop on Benchmarks for Deep Learning in Robotic
  Vision}, 2018.

\bibitem{perazzi17}
Federico Perazzi, Anna Khoreva, Rodrigo Benenson, Bernt Schiele, and Alexander
  Sorkine-Hornung.
\newblock Learning video object segmentation from static images.
\newblock In {\em CVPR}, 2017.

\bibitem{Pfitzner2015Libra3DBW}
Christian Pfitzner, Stefan May, Christian Merkl, Lorenz Breuer, Martin
  Kohrmann, Joel Braun, Franz Dirauf, and Andreas N{\"u}chter.
\newblock Libra3d: Body weight estimation for emergency patients in clinical
  environments with a 3d structured light sensor.
\newblock In {\em ICRA}, 2015.

\bibitem{pinto16}
Lerrel Pinto and Abhinav Gupta.
\newblock Supersizing self-supervision: Learning to grasp from 50k tries and
  700 robot hours.
\newblock In {\em ICRA}, 2016.

\bibitem{fasterrcnn}
Shaoqing Ren, Kaiming He, Ross Girshick, and Jian Sun.
\newblock Faster r-cnn: Towards real-time object detection with region proposal
  networks.
\newblock In {\em NeurIPS}, 2015.

\bibitem{unet}
Olaf Ronneberger, Philipp Fischer, and Thomas Brox.
\newblock U-net: Convolutional networks for biomedical image segmentation.
\newblock In {\em MICCAI}, 2015.

\bibitem{rubinstein13}
Michael Rubinstein, Armand Joulin, Johannes Kopf, and Ce Liu.
\newblock Unsupervised joint object discovery and segmentation in internet
  images.
\newblock In {\em CVPR}, 2013.

\bibitem{schaul16}
Tom Schaul, John Quan, Ioannis Antonoglou, and David Silver.
\newblock Prioritized experience replay.
\newblock In {\em ICLR}, 2016.

\bibitem{siva13}
Parthipan Siva, Chris Russell, Tao Xiang, and Lourdes Agapito.
\newblock Looking beyond the image: Unsupervised learning for object saliency
  and detection.
\newblock In {\em CVPR}, 2013.

\bibitem{Standley2017image2massET}
Trevor~Scott Standley, Ozan Sener, Dawn Chen, and Silvio Savarese.
\newblock image2mass: Estimating the mass of an object from its image.
\newblock In {\em CoRL}, 2017.

\bibitem{stretcu15}
Otilia Stretcu and Marius Leordeanu.
\newblock Multiple frames matching for object discovery in video.
\newblock In {\em BMVC}, 2015.

\bibitem{tang18}
Peng Tang, Xinggang Wang, Angtian Wang, Yongluan Yan, Wenyu Liu, Junzhou Huang,
  and Alan~L. Yuille.
\newblock Weakly supervised region proposal network and object detection.
\newblock In {\em ECCV}, 2018.

\bibitem{tokmakov17}
Pavel Tokmakov, Karteek Alahari, and Cordelia Schmid.
\newblock Learning video object segmentation with visual memory.
\newblock In {\em ICCV}, 2017.

\bibitem{tsai16}
Yi-Hsuan Tsai, Ming-Hsuan Yang, and Michael~J. Black.
\newblock Video segmentation via object flow.
\newblock In {\em CVPR}, 2016.

\bibitem{vanHoof14}
Herke van Hoof, Oliver Kroemer, and Jan Peters.
\newblock Probabilistic segmentation and targeted exploration of objects in
  cluttered environments.
\newblock {\em Transactions on Robotics}, 2014.

\bibitem{wang19zero}
Wenguan Wang, Xiankai Lu, Jianbing Shen, David~J. Crandall, and Ling Shao.
\newblock Zero-shot video object segmentation via attentive graph neural
  networks.
\newblock In {\em ICCV}, 2019.

\bibitem{wang19}
Wenguan Wang, Hongmei Song, Shuyang Zhao, Jianbing Shen, Sanyuan Zhao, Steven
  C.~H. Hoi, and Haibin Ling.
\newblock Learning unsupervised video object segmentation through visual
  attention.
\newblock In {\em CVPR}, 2019.

\bibitem{wu15}
Jiajun Wu, Ilker Yildirim, Joseph~J Lim, Bill Freeman, and Josh Tenenbaum.
\newblock Galileo: Perceiving physical object properties by integrating a
  physics engine with deep learning.
\newblock In {\em NeurIPS}, 2015.

\bibitem{wu2019detectron2}
Yuxin Wu, Alexander Kirillov, Francisco Massa, Wan-Yen Lo, and Ross Girshick.
\newblock Detectron2.
\newblock \url{https://github.com/facebookresearch/detectron2}, 2019.

\bibitem{xiao2016}
Fanyi Xiao and Yong~Jae Lee.
\newblock Track and segment: An iterative unsupervised approach for video
  object proposals.
\newblock In {\em CVPR}, 2016.

\bibitem{khz2020visualreaction}
Kuo-Hao Zeng, Roozbeh Mottaghi, Luca Weihs, and Ali Farhadi.
\newblock Visual reaction: Learning to play catch with your drone.
\newblock In {\em CVPR}, 2020.

\bibitem{zitnick14}
Larry Zitnick and Piotr Dollar.
\newblock Edge boxes: Locating object proposals from edges.
\newblock In {\em ECCV}, 2014.

\end{thebibliography}
}
\newpage
\appendix
\section*{Appendix}
\section{Model details}
\label{app:model}
Our backbone begins with a $5\times5$ convolution with stride $3$ and $32$ channels. All convolutions except for the final ones are followed by BatchNorm and ReLU, which we will assume implicitly from now on. The initial convolution is followed by three convolutional blocks of stride 2 with $64$, $128$ and $256$ output channels, respectively. Each block consists of two $3\times3$ convolutions with a $1\times1$ convolution in between, and a residual connection between the input of the block and the input of the last convolution. 

The blocks are followed by three transposed-convolutional modules, with lateral connections to the inputs of the block, inspired by the original UNet. Each of the modules applies a $2\times2 $ transposed convolution with stride 2 to the output of the corresponding block, concatenates the result with the input to the block, and then applies a $3\times3$ convolution to obtain a tensor of the same shape as the input to the block. An exception is the final transpose-convolutional module, where we increase the number of output channels to $64$. 

The resulting $64\times100\times100$ tensor is concatenated with absolute pixel coordinates, and $1\times1$ convolved to obtain the final output of the backbone, a tensor of size $128\times100\times100$.

The computation of each $128$ dimensional entry in this tensor depends on the input inside a $137\times137$ box (at the original $300\times300$ resolution). This relatively small receptive field is still large enough to fit the vast majority of objects in our dataset, and, as typical for UNet-like architectures, our model output also has direct access to intermediate features computed at high resolution, enabling precise localization of features.

The prediction heads are simple $1\times1$ convolutions, with number of channels equal to $1$ for interaction scores, $3$ for force logits, and $16$ for the embedding vectors. In total, the model has 1.4M parameters.

\section{Loss details}
\label{sec:losses}
\textbf{Interaction score loss.}
Denote by $\mathsf{fg}_{\mathbf p}$, $\mathbf p \in [0,\ldots,100]^2$, the \textit{foreground}, constructed by placing $1$s at positions $\mathbf p$ of successful interactions into an initially empty mask, and then convoluting with the $5\times5$ filter
\begin{equation}\label{eq:kernel}
    \mathsf{kernel} = (e^{-u^2 - v^2})_{u,v \in [-2:2]}.
\end{equation}
Similarly, denote by $\mathsf{bg}_{\mathbf p}$ the \textit{background}, constructed in the same way from the unsuccessful interaction locations.

Let $s_{\mathbf p}$ be the pixel-wise interaction score. Then, our \textit{loss gradient} for $s_{\mathbf p} $ is given by 
\begin{equation}
    \mathsf{grad}_{s_{\mathbf p}}  = \mathsf{fg}_{\mathbf p} \cdot \sigma(-s_{\mathbf p}) \cdot e^{-\frac12 (\max(s_{\mathbf p},0))^2} - \mathsf{bg}_{\mathbf p} \cdot \sigma(s_{\mathbf p}) \cdot e^{-\frac12 (\min(s_{\mathbf p},0))^2},
\end{equation}
where $\sigma$ is the sigmoid function. Our loss function is the integral of that gradient (only the gradient itself is needed for training). In the absence of the exponential factors in $\mathsf{grad}_{s_{\mathbf p}} $, this would be a Kullback-Leibler divergence type loss. The exponential factors ensure that confident scores give rise to very small gradients, like in the focal loss, but our suppression of such gradients is more aggressive than for the standard focal loss.

\textbf{Force loss.} 
There is a force magnitude $r$ ($r=0,1,2$) associated to each successful interaction (namely the magnitude predicted at the time of interaction), and feedback reflecting whether this force was: 1. just right, 2. too small, or 3. too large. Form the $3\times100\times100$ tensor  $(\tilde{\mathsf{ft}}_{\mathbf p}^r)_{r,\mathbf p}$ that is nonzero only at pixels $\mathbf p$ of successful interactions, and such that, if $\mathbf p$ is a successful pixel with predicted force $r$, we have  
\begin{equation}
    \tilde{\mathsf{ft}}_{\mathbf p}^{r'} = \left\{\begin{array}{ll} \mathbbm{1}(r'=r) & \mathrm{case}\, 1 \\ \mathbbm{1}(r'<r)&  \mathrm{case}\,2 \\ \mathbbm{1}(r'>r)&  \mathrm{case}\,3  \end{array} \right. 
\end{equation}
Next, normalize $\tilde{\mathsf{ft}}$ to have mean zero over its first dimension (the one of size $3$), and $L^1$ norm equal to $1$ over this dimension wherever it is nonzero. Finally, convolve the result with $\mathsf{kernel}$ from the previous paragraph to obtain $( \mathsf{ft}_{\mathbf p}^r)_{r,\mathbf p}$, the \textit{force targets} for the predicted pixel-wise force logits $( m_{\mathbf p}^r)_{r,\mathbf p} $.

Using this notation, the loss gradient for the force logits $( m_{\mathbf p}^r)_{r,\mathbf p} $ is given by
\begin{equation}
    \mathsf{grad}_{m_{\mathbf p}^r}  =\mathsf{ft}_{\mathbf p}^r  \cdot  \Big(\mathbbm{1}\big[\mathsf{ft}_{\mathbf p}^r > 0\big] \cdot \sigma(-m_{\mathbf p}^r) \cdot e^{-\frac12 (\max(m_{\mathbf p}^r,0))^2}  + \mathbbm{1}\big[\mathsf{ft}_{\mathbf p}^r  < 0\big] \cdot \sigma(m_{\mathbf p}^r) \cdot e^{-\frac12 (\min(m_{\mathbf p}^r,0))^2}\Big). 
\end{equation}
This formula is similar to the one used in the loss of the interaction score and was inspired by the same considerations. We notice considerable noise in the gradients for the force logits, and we address this by adjusting the gradients $\mathsf{grad}_{m_{\mathbf p}^r} $ by a factor of $0.1$ relative to those of the interaction score and embedding loss.

\textbf{Embedding loss.}
A loss gradient is associated to each noisy segmentation mask corresponding to a successful interaction of the agent. The gradients for each such mask are weighted inversely to the area (number of pixels) of the mask and then added, which defines the full loss gradient for the embedding vectors.

We now define the loss gradient for a single noisy segmentation mask $B^+ $. We denote by $e_{\mathbf p} $ the $16\times100\times100 $ tensor of pixel embedding vectors, and introduce the mean embedding vector $e = \frac1{|B^+|} \sum_{\mathbf p\in B^+}e_{\mathbf p}$.

We compute the $100\times 100$ tensor $d_{\mathbf p} = \mathsf{huber\_square}(e_{\mathbf p} - e), $ where $\mathsf{huber\_square}(x) $ computes the $16D$ squared $L^2$ norm in the forward pass, but in the backward pass truncates gradients to have absolute value less than $1$ (we want to avoid larger gradients for stability of training). Loss gradients for $d_{\mathbf p}$ are given by the following formula, and define gradients for the embedding vectors $e_{\mathbf p} $ by backpropagation:
\begin{equation}
    \mathsf{grad}_{d_{\mathbf p}} = \mathbbm{1}(\mathbf p\not\in B^+) \cdot e^{-(d_{\mathbf p} / 1.5)^4} - \mathbbm{1}(\mathbf p\in B^+) \cdot \frac{1.5\cdot d_{\mathbf p}}{ 1 + d_{\mathbf p}} .
\end{equation}
Note that this gradient increases the distances $d_{\mathbf p}$ to the mean vector $e$ for pixels $\mathbf p\not\in B^+$, but decreases the distances for $\mathbf p\in B^+$. The gradient magnitude decreases for confident predictions ($\mathbf p \in B^+ $ and $d_{\mathbf p} $ close to zero, or $\mathbf p\not\in B^+$ and $d_{\mathbf p}$ large). Finally, the factor $1.5$ sets a scale for the distances $d_{\mathbf p}$: We have 
\begin{equation}
    e^{-(d_{\mathbf p} / 1.5)^4} \approx \frac{1.5\cdot d_{\mathbf p}}{ 1 + d_{\mathbf p}} \qquad \mathrm{at} \qquad d_{\mathbf p} = 1.
\end{equation}
In other words, the gradient magnitudes for positive (on mask) and negative (off mask) pixels balance when the distance to mean vector $e$ of such a pixel is equal to $1$. This explains why we use threshold $\Vert e_{\mathbf p} - e\Vert_2 = 1$ in the clustering algorithm. The reasoning is the same as for binary cross entropy, where we threshold logits $x$ at $0$, the point where the gradient for the positive class, $\sigma(-x)$, equals $\sigma(x)$, the gradient for the negative class.

Note that, if perfect convergence could be achieved during training, the choice of the threshold would be irrelevant (namely, $\Vert e_{\mathbf p} - e\Vert_2$ would be equal to zero for $\mathbf p \in B^+$ and infinity for $\mathbf p \not\in B^+$). Because of noise and limited capacity of the model, convergence cannot be achieved, and indeed, we observe that threshold $1$ in the clustering algorithm usually gives the best performance.

\section{Priorities in the memory bank}
\label{app:memory}
Every image in the memory bank has an associated \textit{priority}, and we sample from the memory bank by first normalizing the priorities across the bank and then sampling without replacement from the resulting probability distribution. A new image is added with priority equal to $0.5$. Its priority gets updated each time it is used in a forward pass, as described below:

We associate a mask $\mathsf{pred\_mask} $ predicted by the current model to each noisy segmentation mask $B^+ $ on the image, as follows: Compute the distances $d_{\mathbf p}$ as in Section~\ref{sec:losses} above, and define $\mathsf{pred\_mask}  = d_{\mathbf p} < 1 $ (a binary mask). Now, compute the IoU of $B^+$ with $\mathsf{pred\_mask} $. We define the \textit{score} of the image to be the minimum of these IoUs for all noisy instance masks $B^+$ in the image. In case the image contains no instances, we set the score equal to $0.5$.

The new priority of the image is computed from its score by the formula 
\begin{equation}
    \mathsf{priority}(\mathsf{score}) = (\mathsf{score} - 0.5)^2 + 0.02.
\end{equation}
The priority is thus minimal (although nonzero, thanks to the offset $0.02$) for scores around $0.5$, and maximal for either low or high scores, corresponding to small or large minimal IoUs, respectively. 

Although we do not claim that more intuitive strategies to choose priorities, like only prioritizing images with small IoUs (following the prioritized replay philosophy), or only those with large IoUs (following certain heuristics in the area of self-paced learning), could not be equally beneficial as the strategy chosen here, we were unable to successfully apply such intuitive strategies in preliminary experiments. We have observed qualitatively that the IoU of noisy instance masks with the \textit{ground truth} mask of the corresponding object is often around $0.5$, and based on this observation, it could be conjectured that sampling images with such noisy masks less often exposes the model to a cleaner signal for instance segmentation.

Note that we do not perform a bias correction of loss gradients, as is sometimes done to ensure the loss gradient remains unbiased despite non-uniform sampling. We have found this to be detrimental to learning in our case.

\section{Implementation details}
\label{sec:implementation}

\textbf{Dataset and interaction.} To initialize a scene for our agent, the \thor~controller requires the position (2 coordinates) and direction of view (2 angles) of the agent within the scene, as well as a random seed for spawning objects into the scene. We choose these parameters from a fixed dataset in order to reduce some of the randomness in training.

The agent interacts with a scene only by applying forces, and its maximum radius of interaction is limited to 1.5 meters. All moveable objects whose ground truth segmentation mask includes at least 10 pixels that are within this distance of the agent are included in the ground truth.

In \thor, forces of a given magnitude and direction are applied to the selected point for 1 millisecond, and induce realistic rigid body motion. Points are selected by emitting a ray from the camera through the selected pixel. The first point hit by this ray and corresponding to an object or structure is selected, unless it is out of reach of the agent (farther than 1.5 meters from the camera).

\textbf{Training details.}
Our network is randomly initialized, and first trained on the instance segmentation task only. We start by filling the memory bank with interactions from 3000 locations of the train set. We then alternate between adding batches from $70$ locations to the memory bank, and training on $K$ batches of size $64$ sampled from the memory bank, where $K$ is linearly annealed from $15$ to $45$ over the course of the training. At each location, the model performs $N=10$ ``greedy'' interaction attempts according to the algorithm of Section 4.1 of the main text, and an additional $10$ random interactions. Our optimizer is Adam with weight decay $10^{-4}$ and learning rate $ 5\cdot 10^{-4}$. In total, we perform interactions at approximately 65k locations, and roughly 25k gradient steps.

After this segmentation-only pre-training, we train segmentation and relative mass prediction jointly. We use essentially the same setting as before, but anneal $K$ from $15$ to $35$, and only perform $5$ ``greedy'' and $5$ random interaction attempts at each location.

\textbf{Hardware details.}
In the setting above, our model can train on a single GeForce RTX 2080 Ti. Interaction with the environment can be parallelized, and needs additional GPU and especially CPU resources. On our 36 CPU Intel Core i9-7980XE machine, we achieved good performance by running 35 agents in parallel, which require an additional RTX 2080 GPU to simulate physics. In total, training takes about 36 hours in this setup. For inference, our current implementation processes 300$\times$300 RGB+D images at about 20-25 fps on a single GPU.


\textbf{Training details -- Mask-RCNN.}
Our Mask-RCNN references are trained in detectron2 \cite{wu2019detectron2} in two settings: fully supervised (all model inputs pre-rendered, and learning targets use full ground truth information), and instance segmentation with oracle interactions (target segmentation masks are replaced by the output of our self-supervision module). In the latter case, data collection is still independent of the model, and is done prior to training. Training itself can then follow the supervised paradigm (but with noisy targets).

The model setup follows a standard configuration for Mask-RCNN with a ResNet-18 backbone network (all defaults as in detectron2), but with RGB+D input channels. Networks are randomly initialized and trained from scratch. We use a base learning rate of $2\cdot10^{-3}$ and reduce it by a factor $10^{-1}$ after 60,000 and 80,000 steps, before terminating at 90,000 steps.

\textbf{Providing more supervision.} We describe the interaction and mask oracles used to study the effects of partial supervision on the model performance.

The interaction oracle first filters the actions predicted by the model to contain at most one point on each moveable object that is in reach. It then adds one such point for each object that was not selected for interaction by the model. Other selected points are not affected by the oracle, so that the model can produce its own hard negatives. This modified list of interaction points is then passed on to the agent for interaction.

For each interaction point (2-D), the mask oracle retrieves the ground-truth segmentation mask that point belongs to, or returns the empty mask if the point either does not belong to an object or corresponds to a world location not in reach of the agent. These masks are down-sampled and substituted for the ones usually produced by interaction and the self-supervision module. The rest of the training remains identical to the self-supervised case.

\textbf{The videoPCA baseline.} 
The method \cite{croitoru17} trains a foreground-background segmentation model via $L^2$ loss to a soft, noisy segmentation mask extracted from videos. At inference, the foreground is clustered into connected components (instances), and post-processed. We adapted this approach to be applicable to our dataset in the following way:

The video is obtained from \thor~by performing oracle interactions as described above. These videos are much simpler than the ones for which the original approach of \cite{croitoru17} was designed, and we use less (namely, 4) PCA components for frame reconstruction. Also, moving objects are often not in the center of the image. We replace the centering heuristic of \cite{croitoru17} by centering around the hard masks produced in the first step of their approach. Soft masks corresponding to different interactions are added and clipped at $1$ to give a single soft mask target. Finally, we adapt the sizes of filters they use for smoothing to the resolution of our videos.

The model we use has the same backbone as our self-supervised model, and is trained for the same number of steps and with the same learning rate, but using $L^2$ loss to a single soft mask score, as in \cite{croitoru17}. For post-processing, we use superpixels (rather than a CRF, as in the original).

\textbf{Optical flow.}
Unsupervised optical flow is a principled approach to finding correspondences between images before and after interaction, and thresholding the optical flow in the initial image can ideally provide clean segmentation masks of the moving object. However, learning unsupervised optical flow is not straightforward and itself prone to noise. In preliminary experiments, we used the pretrained (unsupervised) Unflow network from \cite{meister18} to extract masks in this way, but did not achieve good performance. It is an interesting research direction to train an optical flow network on our self-supervised data from scratch.

\textbf{Ablation with $\infty$ arm length.}
We have performed various experiments investigating the effect of the interaction radius of the agent on training. An interesting case is infinite interaction range, where the agent can apply forces to all objects in sight, irrespective of their distance to the agent. In Table 1, row (l), of the main text, we report the performance of such an agent. 

\textbf{Selection of hyperparameters.} The model, losses, and self-supervision module incorporate a large variety of design and hyperparameter choices. Most of these choices were made using domain knowledge, and were not tuned or ablated by evaluation on the ground truth. There are two exceptions:
\begin{itemize}
    \item We studied four settings for the interaction range of the agent. In \thor, this range can be a sphere around the camera, a vertical cylinder around the agent, or an intersection of the two. A larger interaction range means more potential signal for the agent to learn from, but also more ground truth objects for the agent to discover. Among the settings we tried, the one we report on (sphere of radius 1.5m) was the second best performing. This is a natural setting (robot arm anchored at a fixed point on the robot).
    \item We studied five choices for the filter $\mathsf{kernel}$ defined in (\ref{eq:kernel}), and the threshold $1.5$ defined in Section~\ref{sec:selfsupmodule} below. We found it to be important that this filter be small and its elements quickly decaying. Especially in its use in the self-supervision module, large filter sizes have a surprisingly detrimental effect. We used the best performing choice in our final setup.
\end{itemize}

\section{Self-supervision module details}
\label{sec:selfsupmodule}
\textbf{Successful vs. unsuccessful interactions.}
Every interaction produces a $100\times100$ binary mask $B^+$ via thresholding of the difference of images before and after interaction, and alignment with superpixels. Such masks contain various sources of noise, and can be non-empty even if the interaction does not actually move any object in the scene. It is effective to declare interactions unsuccessful if they satisfy the following criterion: restrict the mask to the $5\times5$ neighborhood of the point of interaction, and multiply with the filter $\mathsf{kernel}$ from equation  (\ref{eq:kernel}); then, the interaction is unsuccessful if the sum of the entries of the resulting $5\times5$ matrix is less than the threshold $1.5$. Otherwise, the interaction is successful.

Intuitively, successful interactions thus produce masks that contain at least two or three $3\times3$ grid cells (at full resolution) in the immediate neighborhood of the point of interaction (with more weight given to cells close to the point).

\section{Details on results}
\begin{figure}[tp]
    \centering
    \includegraphics[width=33pc]{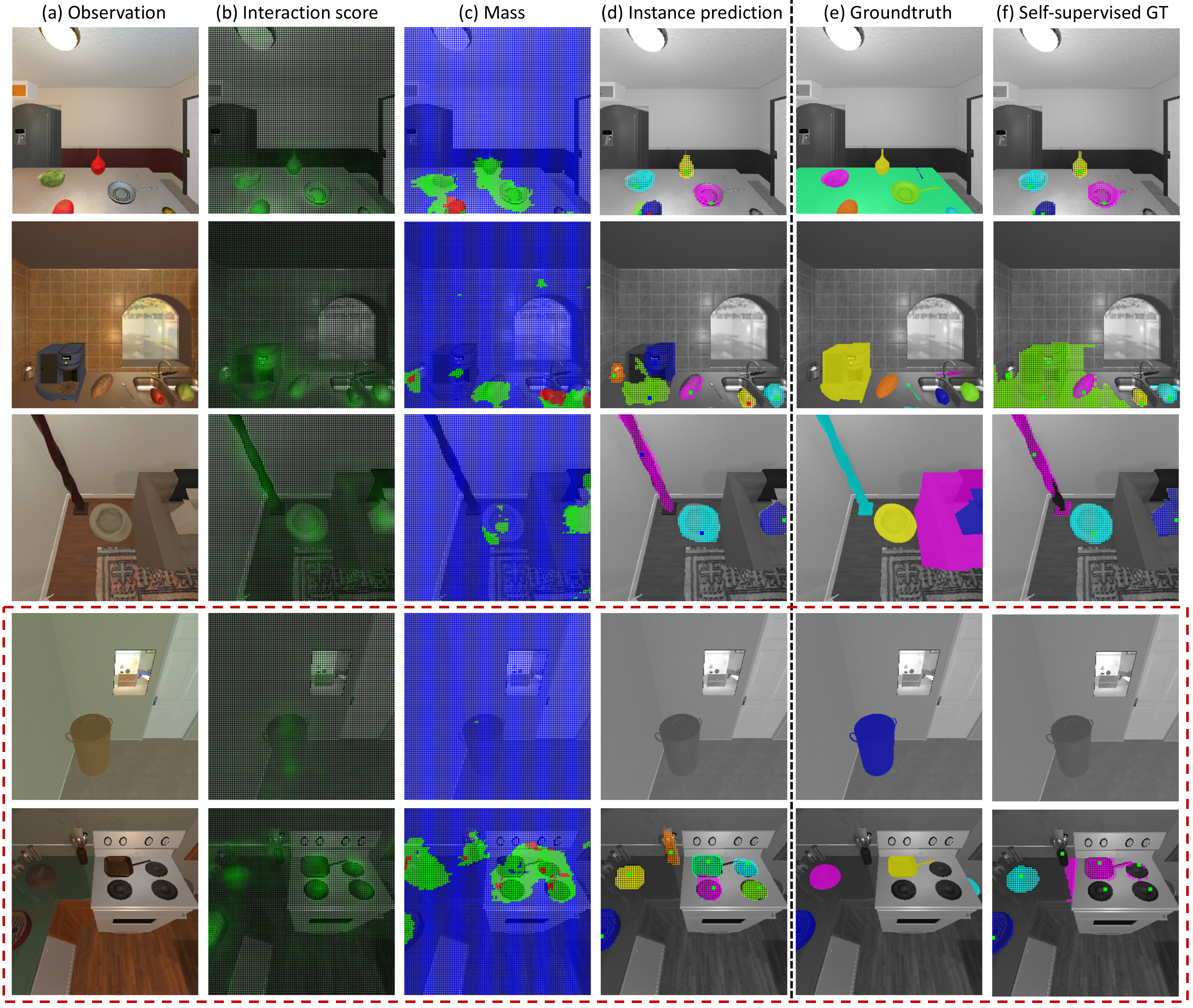}
    \caption{\textbf{Additional qualitative results.} (a) Initial observation. (b) Predicted interaction scores. The brighter green corresponds to higher interaction score. (c) Predicted mass. The masses are shown in red (light), green (medium) and blue (heavy). (d) Instance prediction results. The selected interaction points are shown with colored dots. (e) The groundtruth that for objects in the vicinity of the agent. (f) Self-supervised groundtruth, which is noisy and is obtained by interaction. The bottom two rows show failure examples.}
    \label{fig:qual_supp} 
\end{figure}

\textbf{More qualitative examples.} We include some additional qualitative examples in Figure~\ref{fig:qual_supp}. They serve to illustrate the following observations:
\begin{itemize}
    \item As seen in rows 1 to 3, our model can learn to detect multiple objects of varying shapes, even in relatively cluttered scenes. It does so by learning from a self-supervised ground truth that is sometimes very noisy (as in row 2), and is usually missing objects (like the table and sofa in rows 1 and 3, which are hard to move, and several small objects, which are hard to localize). Relative mass predictions reliably single out heavy from medium or light objects, and the prediction is not just based on the size of the object proposal (e.g. the cardboard box in row 3 is large, but can easily be moved by a medium force).
    \item Rows 1 to 3 also contain some common failure modes of our model, like over-segmentation, or including objects that are just out of reach (the bottle in the left corner of the images in row 2).
    \item As seen in row 4, our model sometimes lacks confidence in selecting even easy to move, large objects in uncluttered scenes. 
    \item Conversely, as seen in row 5, our model sometimes confidently selects parts of structures as objects. Note that some of the false positives in that image (especially the kitchen utensils to the left of the stove, which are not moveable and part of a structure in \thor) could well be a moveable object if judged solely by visual clues. Interaction is necessary to learn that they are not.
\end{itemize}

\textbf{The variance of the results.}
Since our model creates its own learning signal via self-supervision, final performance of training could be expected to be more sensitive to random seeds than is usually the case in fully supervised training. However, we have not observed large fluctuations as appear sometimes in reinforcement learning. Training is too expensive for us to collect standard deviations for all results reported in the main paper. Instead, we performed three additional self-supervised training runs on the \textit{NovelObjects} dataset, and report mean and standard deviation among the runs on this dataset in Table~\ref{tab:std}. This serves to illustrate the fluctuations in performance that can be expected for our self-supervised approach.

\textbf{Confusion matrix for mass estimation.} The Mean per Class Accuracy is only one measure to evaluate the performance of our model on the relative mass prediction task. More detail is contained in the confusion matrix of relative mass predictions among successfully detected objects (as in the main text, the BBox-IoU threshold for a successful detection is $0.5$). The confusion matrices on both test sets can be found in Table~\ref{tab:confmat}. Their rows are normalized. Mean per Class Accuracy is the average of the diagonal entries of the confusion matrix.

\begin{table*}[t!]\footnotesize
  \centering
\begin{small}
\centerline{
    \begin{tabular}{cc cc cc}
    \toprule
    \multicolumn{2}{c}{\centering{Bounding Box}} & \multicolumn{2}{c}{\centering{Mask}} & \multicolumn{2}{c}{\centering{Mass estimation}} \\
        AP$^{\scriptsize{\mbox{IOU=0.5}}}$ & AP & AP$^{\scriptsize{\mbox{IOU=0.5}}}$ & AP& 
        Mean per Class Acc. & AP$^{\scriptsize{\mbox{IOU=0.5}}}$ (Mass \& BBox)  \\
    \midrule
    $23.77\pm 0.83$ & $11.33\pm 0.59$ & $21.92\pm 0.90$ & $10.01\pm 0.50$ & $49.79\pm 1.36$ & $10.75\pm 1.03$ \\
    \midrule
    \multicolumn{6}{l}{\bf Run reported in main text (for reference):} \\
    $24.19$ & $11.65$  & $22.00$ & $10.24$ & $50.79$ & $11.85$ \\
    \bottomrule
  \end{tabular}
  }
  \end{small}
  \caption{\textbf{Stochasticity of model performance.} Results are for \emph{NovelObjects} scenarios.}
  \label{tab:std}
\end{table*}

\begin{table*}[t!]\footnotesize
  \centering
\begin{small}
\centerline{
    \begin{tabular}{c c}
    \toprule
    \textit{NovelObjects} & \textit{NovelSpaces} \\
    \midrule
    $\left[ \begin{array}{ccc} 0.1318 & 0.5041 & 0.3641 \\ 0.0510 & 0.6114 & 0.3376 \\ 0.0066 & 0.2128 & 0.7805 \end{array}\right]$ & $  \left[ \begin{array}{ccc} 0.4193 & 0.5092 & 0.0714  \\ 0.1591 & 0.5413 & 0.2996 \\ 0.0366 & 0.2483 & 0.7151 \end{array}\right]$ 
  \end{tabular}
  }
  \end{small}
  \caption{\textbf{Confusion matrices for mass estimation.} Rows index ground truth classes and columns predicted classes. Indices $0,\,1,\,2$ denote ``light'', ``medium'', ``heavy''. Computed for successful detections in the corresponding test set (with BBox - IoU threshold of $0.5$).}
  \vspace{4cm}
  \label{tab:confmat}
\end{table*}

\end{document}